# Tackling Tuberculosis: A Comparative Dive into Machine Learning for Tuberculosis Detection

By Daanish Hindustani[1], Sanober Hindustani[2], Preston Nguyen[1]

[1] College of Science and Engineering, University of Minnesota –Twin Cities, [2] College of Biological Sciences, University of Minnesota –Twin Cities

**Abstract:** This study explores the application of machine learning models, specifically a pretrained ResNet-50 model and a general SqueezeNet model, in diagnosing tuberculosis (TB) using chest X-ray images. TB, a persistent infectious disease affecting humanity for millennia, poses challenges in diagnosis, especially in resource-limited settings. Traditional methods, such as sputum smear microscopy and culture, are inefficient, prompting the exploration of advanced technologies like deep learning and computer vision.
The study utilized a dataset from Kaggle, consisting of 4,200 chest X-rays, to develop and compare the performance of the two machine learning models. Preprocessing involved data splitting, augmentation, and resizing to enhance training efficiency. Evaluation metrics, including accuracy, precision, recall, and confusion matrix, were employed to assess model performance. Results showcase that the SqueezeNet achieved a loss of 32%, accuracy of 89%, precision of 98%, recall of 80%, and an F1 score of 87%. In contrast, the ResNet-50 model exhibited a loss of 54%, accuracy of 73%, precision of 88%, recall of 52%, and an F1 score of 65%. This study emphasizes the potential of machine learning in TB detection and possible implications for early identification and treatment initiation. The possibility of integrating such models into mobile devices expands their utility in areas lacking TB detection resources. However, despite promising results, the need for continued development of faster, smaller, and more accurate TB detection models remains crucial in contributing to the global efforts in combating TB.

## Introduction

Tuberculosis (TB) is an infectious disease that has been impacting humanity for the past 4,000 years [5]. This chronic illness is caused by the bacillus Mycobacterium tuberculosis. TB spreads through the air, making it a threat in highly dense areas [5] [4]. As of 2007, there have been an estimated 9.27 million incident cases of TB, 13.7 million prevalent cases, and 1.32 million deaths from TB globally [5]. Eighty-six percent of all cases originate from Asia and Africa alone. The lack of needed resources for both detection and treatment, as well as highly densely populated environments, leads to many outbreaks and deaths [6].

The main strategy for diagnosing TB includes sputum smear microscopy and culture. However, due to the time and resources needed to utilize these strategies, it is very inefficient, especially in third-world countries where supplies are limited [6]. Currently, X-rays are being utilized to diagnose TB; however, this method also has its issues. The first issue is time; the amount of time a doctor might take to analyze and determine if TB is present takes away from treating the patient. The second issue is accuracy; although these doctors have years of experience and education, mistakes can still occur, which is detrimental to the patient's health [6].

Advancements in deep learning (DL) and computer vision (CV) have led to a better understanding of TB diagnosis, providing more efficient and accurate alternatives to traditional methods of detection. Deep learning algorithms can be trained on large datasets of chest X-rays and can quickly and accurately identify patterns associated with TB infections. This approach not only reduces the time required for diagnosis but also makes them valuable tools in resource-limited settings. In this study, we sought to compare two different machine learning models in diagnosing TB using X-rays. The purpose of this was to evaluate the practicality of these models in the real world and to assess the speed and accuracy of these model. With limited resources around the world, it's important to explore more cost-effective diagnostic techniques to improve the overall health of underdeveloped regions.

**Related Research**

Machine learning (ML) has been extensively used in the medical industry, ranging from detecting to diagnosing various ailments. The Conference on Trends in Electronics and Informatics introduced four computer vision models (CV) that enable fast detection of brain tumors from MRI images[2]. These models consisted of a CRF (Conditional Random Field), an SVM (Support Vector Machine), a CNN (Convolutional Neural Network), and a GA (Genetic Algorithm). The resulting accuracy and efficiency were 89% and 87.5% for the CRF, 84.5% and 90.3% for the SVM, 91% and 92.1% for the CNN, and 91% and 92.7% for the GA[2]. This study showcased the potential of utilizing CV models in detecting cancers earlier, ensuring expedited treatment for the patient and increasing their odds of survival.

The International Conference on Smart Electronics and Communication (ICOSEC) introduced a CNN model for Alzheimer's disease (AD) detection[8]. The team's CV model was able to classify AD using MRI images. The dataset included three classes of images with a total number of 1512 mild, 2633 normal, and 2480 AD cases[8]. An accuracy of 99% was achieved, showcasing the potential of CV in detecting AD earlier and potentially treating the patients quickly to minimize brain damage[8].

These research projects indicated the potential of CV in the medical industry and, more importantly, the potential for aiding practitioners in the detection and treatment of many diseases earlier.

**Convolution Neural Networks**

A general understanding of Convolutional Neural Networks (CNN) is needed to grasp the true potential of ML in the healthcare industry. Convolutional Neural Networks (CNNs) are a type of deep learning model specifically designed for processing images. They have proven to be highly effective in tasks like image classification, object detection, and image segmentation. The core building blocks of CNNs are the convolutional layers. These layers use small filters (also called kernels) that slide or convolve across the input image to extract local patterns or features. Each filter detects specific features like edges, textures, or more complex structures. Multiple filters are used in parallel to capture different aspects of the input. This is what makes CNNs particularly impactful for classification tasks like tumor detection, AD detection, and even TB detection. The ability to hone in on a particular segmentation of the image while ignoring background noise allows for innate and accurate detections of medical ailments.

An activation function (ReLU) is applied to each convolution as well as a pooling layer. This allows for the convolution matrix to be processed even more and hone in on a specific feature of an image like a tumor or TB while also minimizing background noise. Finally, the resulting 3D matrix is flattened and fed into a traditional Neural Network. This allows for backpropagation and in-depth learning of images to occur, increasing classification performance.

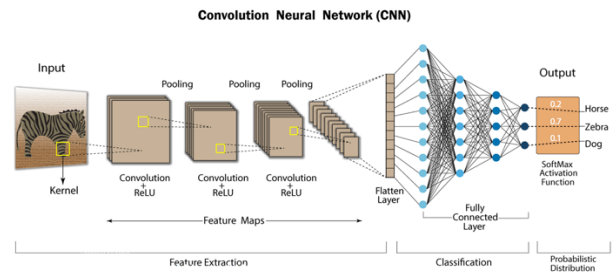

*Figure 1*. Convolutional Neural Network — CNN architecture [9]

**Model Architectures**

This paper outlines two different architectures: a SqueezeNet (SN) and a ResNet-50 (RS50) architecture. An understanding of both architectures, as well as the benefits and potential applications of each, is necessary to ensure the proper integration of these architectures in the medical field. The context behind choosing these two models in this experiment includes establishing a difference between a relatively old model (RS50) and a newer model (SN). This difference was used to showcase the continuous improvement of CV models and the potential for future enhancements of newer models like the SN model.

**SqueezeNet Architecture**

The overall SN architecture is similar to a traditional CNN architecture but with one modification to the convolution layer. The resulting convolution layer seen in the CNN is replaced with a fire module for the SN convolution layer. This module consists of a squeeze layer followed by an expand layer. The squeeze layer primarily uses 1x1 convolutions to reduce the number of input channels (squeeze), while the expand layer uses a combination of 1x1 and 3x3 convolutions to increase the number of channels (expand)[3]. Using a 1x1 convolution (squeeze layer) is computationally less expensive compared to larger convolutions. Additionally, 1x1 convolutions also reduce the number of input channels and, subsequently, the number of parameters in the network, making the overall model smaller[3]. The expand layer follows the squeeze layer and aims to increase the number of channels in the feature representation.

SqueezeNet offers superior computational efficiency and a reduced model size through its innovative use of 1x1 convolutions and fire modules, making it well-suited for resource-constrained devices. The compact architecture maintains competitive accuracy in image classification tasks, providing a balance between model performance and deployment feasibility. SqueezeNet's strategic design, including global average pooling and bypass connections, enhances both energy efficiency and real-time inference capabilities, contributing to its versatility.

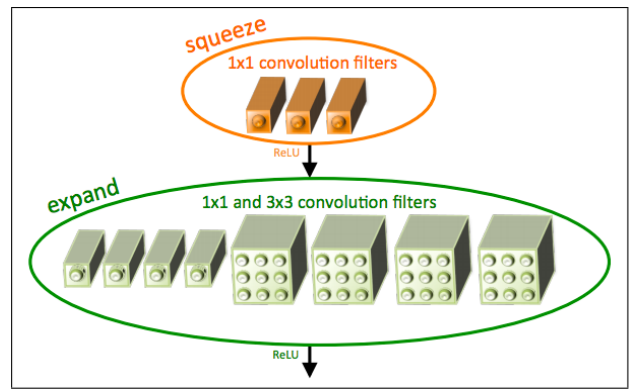

*Figure 2*. Architecture of fire model [3]

**Resnet50 Architecture**

The overall architecture of ResNet-50 (RS50) is similar to the CNN architecture but with a few additional add-ons to the convolution layers. The initial layers consist of standard convolutional layers that extract low-level features from the input image. The subsequent layers are the core innovation of RS50: the residual block. Instead of simply stacking layers one after another, a residual block includes a shortcut connection that skips one or more layers[10]. This shortcut connection allows the network to learn residuals or changes to the input, making it easier to train very deep networks. The following layers follow the same structure as SN and CNN: pooling, flattening, and dense layers[10]. The innovation of the residual block was the first of its kind and revolutionized the way CV models are built today.

ResNet-50's skip connections enable effective training of deep neural networks, addressing the vanishing gradient problem and allowing for the capture of complex hierarchical features[10]. This results in higher accuracy and faster convergence during training. The architecture's versatility, transfer learning capabilities, and efficient parameter learning make it a widely adopted choice for various computer vision tasks, consistently delivering state-of-the-art performance.

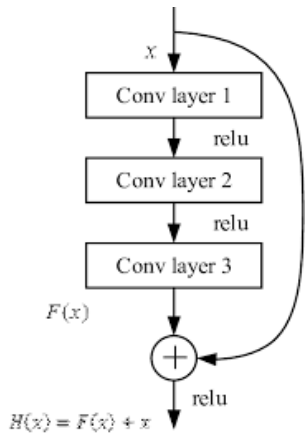

*Figure 3.* Resnet50 residual block [7]

**Image Processing**

This study aimed to develop a tuberculosis (TB) detection model utilizing chest X-rays obtained from datasets on Kaggle.com. The dataset comprised 4200 chest X-rays with dimensions of 512x512x3(pixel*pixel*filter), consisting of 3500 normal chest X-rays and 700 TB-positive chest X-rays for model development. The images were in digital PNG format, RGB filters, and labeled accordingly.

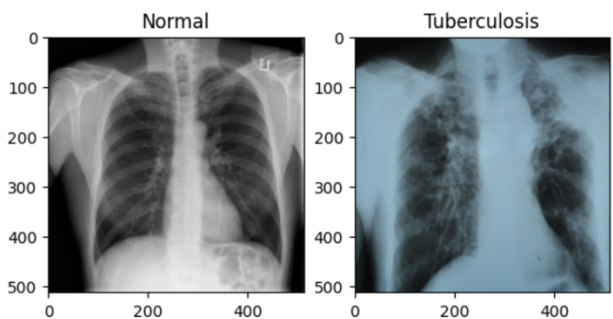

*Figure 4.* Example of the original TB and Normal X-rays.

The initial step in preprocessing these X-rays involved balancing the dataset. This is to ensure a model that can generalize to both image types, not just one. The initial dataset was split into three categories: testing, validation, and training. Six hundred TB images and 600 non-TB images were used in training to achieve proper generalization. Then, 20% of the training data was used as validation data. Finally, 100 TB images and 101 non-TB images were used for testing.

These specific splits were used due to the imbalance of TB-positive and non-TB images. The training splits allow for proper generalization and prevent the model from generalizing to one type of image. The testing split allows for an even number of TB and non-TB images to be tested, allowing for the model to be evaluated properly. The corresponding testing and training/validation data were organized into two directories, with TB and normal data separated within each directory.

The second step in preprocessing the images included augmenting the training data. Image augmentation aids in improving the performance of classification models by exposing them to a broader range of variations during training. For example, rotating and flipping images helps the model learn features from different perspectives, making it more adept at recognizing objects from various angles. Translating and scaling contribute to the model's ability to handle variations in object size and position, enhancing its spatial understanding. Introducing color variations and noise enables the model to adapt to diverse lighting conditions and real-world imperfections, making it more robust and less prone to overfitting on specific patterns from the training data. Overall, image augmentation serves as a regularization technique, fostering a more generalized and resilient model capable of achieving better performance on unseen data.

The training data underwent several augmentation techniques to enhance the robustness and generalization of the classification model. Firstly, the rescaling of the original image by 1/255 normalized the pixel values, allowing for efficient and fast training. Rotation within a range of 40 degrees introduced diversity by randomly rotating images, exposing the model to variations in object orientation. Horizontal and vertical shifting within a range of 0.2(scaling factor) simulated changes in object position, contributing to the model's ability to recognize objects in different spatial locations. Shearing within a range of 0.2(scaling factor) introduced distortions that mimic changes in perspective, aiding the model in handling variations in object shape. Zooming within a range of 0.2(scaling factor) provided exposure to different object sizes during training, which is crucial for scenarios with varying scales. The application of horizontal flipping added mirror variations, making the model invariant to left-right orientation changes. The testing and validation data were rescaled by 1/255 but not augmented, ensuring a fair evaluation of model performance on unaltered images, and providing a realistic measure of its generalization capabilities.

The final preprocessing step involved resizing all images from a 512x512x3(pixel*pixel*filter) format to a 64x64x3(pixel*pixel*filter) format to expedite training speed. Additionally, all images were randomized within their respective directories for training, validation, and testing. The model was crafted in the Kaggle IDE using TensorFlow along with various Python libraries.

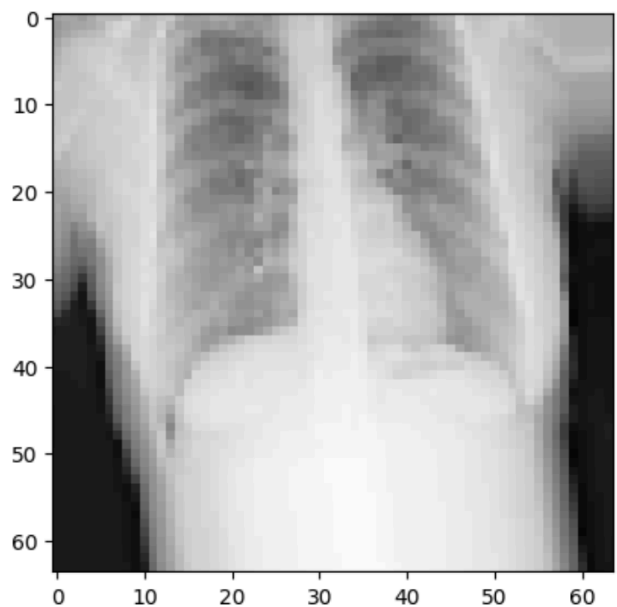

*Figure 5*. Example of the augmented training data.

**Results**

To evaluate the models, four key metrics were employed: accuracy, precision, recall, and F1 score. These metrics were chosen to comprehensively assess the model's performance on unseen data. Specifically, precision and recall offered insights into the model's classification performance on the testing set. Precision gauged the accuracy of positive predictions made by the model, while recall measured the model's ability to correctly identify all relevant instances of a positive class. Additionally, the F1 score amalgamated precision and recall into a single value, providing a well-balanced measure of the model's overall performance. In summary, these metrics provided a comprehensive understanding of the model's strength in tackling the TB classification task.

| Metrics | ResNet50 | SqueezeNet |
|---|---|---|
| F1 Score | 65% | 87% |
| Accuracy | 73% | 89% |
| Loss | 54% | 32% |
| Precision | 88% | 98% |
| Recall | 52% | 80% |

*Table 1.* Resulting metrics of ResNet50 and SqueezeNet model from testing data.

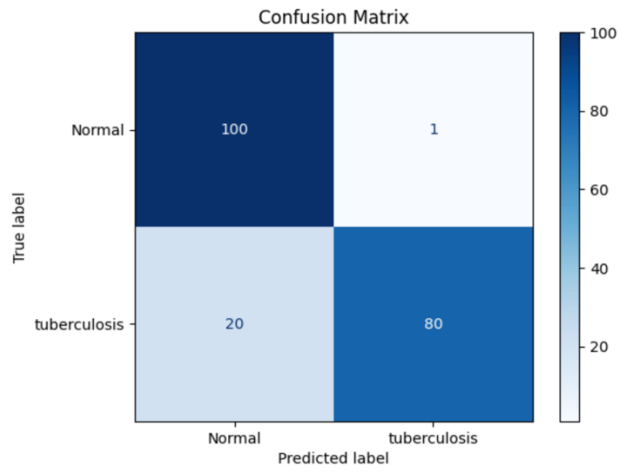

*Figure 6.* Confusion Matrix of the testing data from the SqueezeNet model.

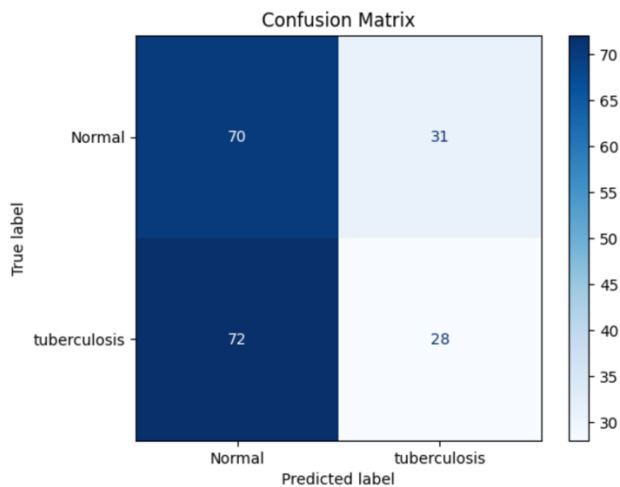

*Figure 7.* Confusion Matrix of the testing data from the ResNet50 model.

**Conclusion**

This study underscores the significant potential of machine learning in revolutionizing the detection of Tuberculosis (TB) through the analysis of chest X-ray images. The two models employed, ResNet50 and SqueezeNet, exhibited varying degrees of performance, highlighting the importance of choosing the appropriate model for specific use cases.

The SqueezeNet model demonstrated superior metrics, with a loss of 32%, an accuracy of 89%, precision of 98%, recall of 80%, and an F1 score of 87%, surpassing the performance of the ResNet50 model, which had a loss of 54%, an accuracy of 73%, precision of 88%, recall of 52%, and an F1 score of 65%. This suggests that SqueezeNet could be a more efficient choice for TB detection, especially in resource-constrained settings.

The SqueezeNet model's lower memory usage compared to ResNet50 makes it a promising candidate for deployment on mobile devices, such as phones. This characteristic extends the reach of TB detection capabilities to regions with limited access to traditional diagnostic resources, presenting a valuable opportunity for early identification and timely initiation of treatment.

While the outcome of this study is promising, it is crucial to recognize the ongoing need for advancements in TB detection models. Future developments should focus on creating faster, smaller, and even more accurate models to ensure precise and efficient TB detection. By addressing these challenges, the integration of machine learning models into clinical practice can significantly contribute to global efforts in combating tuberculosis and implementing effective control strategies.

**References**

1. Dye, C. (2006a). *Global Epidemiology of Tuberculosis*. The Lancet, 367(9514), 938–940. https://doi.org/10.1016/s0140-6736(06)68384-0

2. Hemanth, G., Janardhan, M., & Sujihelen, L. (2019). *Design and implementing brain tumor detection using machine learning approach. 2019 3rd International Conference on Trends in Electronics and Informatics (ICOEI).* https://doi.org/10.1109/icoei.2019.8862553

3. Iandola, F. N., Han, S., Moskewicz, M. W., Ashraf, K., Dally, W. J., & Keutzer, K. (2016, November 4*). Squeezenet: Alexnet-level accuracy with 50x fewer parameters and <0.5MB model size*. arXiv.org. https://doi.org/10.4855/arXiv.1602.07360

4. Luke, E., Swafford, K., Shirazi, G., & Venketaraman, V. (2022). *TB and Covid-19: An exploration of the characteristics and resulting complications of co-infection. Frontiers in Bioscience-Scholar,* 14(1), 6. https://doi.org/10.31083/j.fbs1401006

5. Pai, M., Kasaeva, T., & Swaminathan, S. (2022a*). Covid-19's devastating effect on tuberculosis care — a path to recovery. New England Journal of Medicine*, 386(16), 1490–1493. https://doi.org/10.1056/nejmp2118145

6. Rabahi, M. F., Silva Júnior, J. L. R. da, Ferreira, A. C. G., Tannus-Silva, D. G. S., & Conde, M. B. (2017). *Tuberculosis treatment. Jornal Brasileiro De Pneumologia: Publicacao Oficial Da Sociedade Brasileira De Pneumologia E Tisilogia*, 43(6), 472–486. https://doi.org/10.1590/S1806-37562016000000388

7. Sahoo, S. (2022, September 26). *Residual blocks-building blocks of Resnet. Medium*. https://towardsdatascience.com/residual-blocks-building-blocks-of-resnet-fd90ca15d6ec

8. Salehi, A. W., Baglat, P., Sharma, B. B., Gupta, G., & Upadhya, A. (2020). *A CNN model: Earlier diagnosis and classification of alzheimer disease using MRI. 2020 International Conference on Smart Electronics and Communication (ICOSEC).* https://doi.org/10.1109/icosec49089.2020.9215402.

9. Shahriar, N. (2023, February 1). *What is Convolutional Neural Network-CNN (deep learning). Medium.* https://nafizshahriar.medium.com/what-is-convolutional-neural-network-cnn-deep-learning-b3921bdd82d5

10. Visin, F., Kastner, K., Cho, K., Matteucci, M., Courville, A., & Bengio, Y. (2015, July 23). *Renet: A recurrent neural network based alternative to Convolutional Networks.* arXiv.org. https://doi.org/10.48550/arXiv.1505.00393